\documentclass[lettersize,journal]{IEEEtran}
\usepackage{amsmath,amsfonts}
\usepackage{algorithmic}
\usepackage{algorithm}
\usepackage{array}
\usepackage[caption=false,font=normalsize,labelfont=sf,textfont=sf]{subfig}
\usepackage{textcomp}
\usepackage{stfloats}
\usepackage{url}
\usepackage{verbatim}
\usepackage{graphicx}
\usepackage{cite}
\usepackage{hyperref}
\usepackage{multirow}
\usepackage{booktabs}
\hypersetup{
    colorlinks=true,
    linkcolor=blue,
    filecolor=magenta,      
    urlcolor=cyan,
    citecolor=blue, 
    pdfstartview=FitV,
    pdftitle={My Title},
    pdfauthor={Author},
    pdfsubject={Subject},
    pdfkeywords={keyword1, keyword2, keyword3},
    pdfpagemode=None,
    pdfnewwindow=true 
}
\usepackage{amsmath,amsfonts}
\usepackage{algorithmic}
\usepackage{algorithm}
\usepackage{array}
\usepackage[caption=false,font=normalsize,labelfont=sf,textfont=sf]{subfig}
\usepackage{textcomp}
\usepackage{stfloats}
\usepackage{url}
\usepackage{verbatim}
\usepackage{graphicx}
\usepackage{cite}
\usepackage{booktabs}
\usepackage{color}
\usepackage{makecell}
\begin{document}

\title{A New Type of Adversarial Examples}


\author{Xingyang Nie, Caoliang Zhang, Su Pan, Biao Wang, Huilin Ge, and Tao Fang
\thanks{Xingyang Nie, Caoliang Zhang, Su Pan, Biao Wang, Huilin Ge, and Tao Fang are with the Ocean College, Jiangsu University of Science and Technology, Zhenjiang 212003, China. E-mail: 122386231@qq.com; 241112201127@stu.just.edu.cn; 1242425221@qq.com; wbiaoang@just.edu.cn; gh1989@just.edu.cn; 1007629788@qq.com.}

\thanks{(Corresponding author: Xingyang Nie).}}




\maketitle

\begin{abstract}
Most machine learning models are vulnerable to adversarial examples, which poses security concerns on these models. Adversarial examples are crafted by applying subtle but intentionally worst-case modifications to examples from the dataset, leading the model to output a different answer from the original example. In this paper, adversarial examples are formed in an exactly opposite manner, which are significantly different from the original examples but result in the same answer. We propose a novel set of algorithms to produce such adversarial examples, including the negative iterative fast gradient sign method (negI-FGSM) and the negative iterative fast gradient method (negI-FGM), along with their momentum variants: the negative momentum iterative fast gradient sign method (negMI-FGSM) and the negative momentum iterative fast gradient method (negMI-FGM). Adversarial examples constructed by these methods could be used to perform an attack on machine learning systems in certain occasions. Moreover, our results show that the adversarial examples are not merely distributed in the neighbourhood of the examples from the dataset; instead, they are distributed extensively in the sample space.
\end{abstract}


\section{Introduction}
\IEEEPARstart{M}{achine} learning models, including deep neural networks (DNNs), are often vulnerable to adversarial examples\cite{szegedy2014intriguing,goodfellow2015explaining}. Adversarial examples are maliciously perturbed inputs constructed by adding human-imperceptible noise to examples from the dataset, but mislead a model to incorrect predictions at test time\cite{akhtar2018threat,yuan2019adversarial}.

If the architecture and weights of a model are known, adversarial examples can be constructed in the white-box manner\cite{yang2024black}. The fast gradient sign method (FGSM)\cite{goodfellow2015explaining} and its iterative variant (I-FGSM)\cite{kurakin2017adversarial} are two representative ones among these white-box methods.
In many cases, the adversarial examples designed to be misclassified by one model are still misclassified by others\cite{szegedy2014intriguing,liu2017delving,moosavi2017universal}. The good transferability property of adversarial examples makes black-box attacks possible\cite{papernot2016practical,papernot2016transferability,peng2026c} and poses real security threats since the attacker usually has no access to the underlying model in practice\cite{zhu2025high}.

To illustrate how adversarial examples make a DNN-based system vulnerable and then pose security issues, we set the application scenario as autonomous driving\cite{he2022robust}. Autonomous driving is obviously a safety-critical task. DNNs are now commonly employed in autonomous driving systems to recognize vehicles or traffic signs on the road \cite{li2020adaptive}. Fig. 1a and Fig. 1b are two input images to the trained DNN used in an autonomous driving system. Fig. 1b is an adversarial example generated from Fig. 1a. Fig. 1a is correctly classified as a car, while Fig. 1b is misclassified by the DNN. Altering the car's body as Fig. 1b, though the perturbation is imperceptible, prevents the DNN from recognizing it as a moving vehicle\cite{zhang2019stochastic,hashemi2022improving}. Then, the autonomous driving system will possibly not take proper reaction to avoid the car and eventually cause an accident. Thus, it is crucial for security sensitive systems incorporating DNNs to defend against adversarial examples\cite{kurakin2018adversarial}.

\begin{figure}[!t]
\centering
{\includegraphics[width=1.5in]{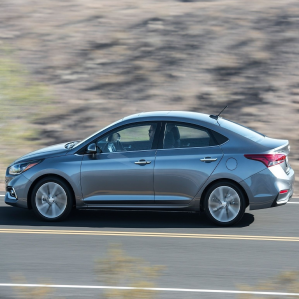}%
\label{fig.1a}}
\hfil
{\includegraphics[width=1.5in]{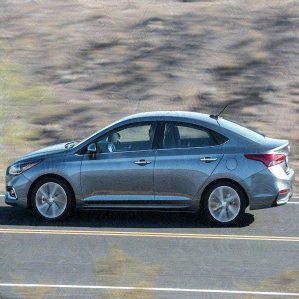}%
\label{fig.1b}}
\caption{Two input images to the DNN used in an autonomous driving system. (a) The original image. (b) The adversarial image generated from (a).}
\label{fig.1}
\end{figure}

Several techniques have been developed to defend against adversarial attacks\cite{zhou2021hierarchical}. Adversarial training is perhaps the most commonly used one among them\cite{ganin2016domain}. Adversarial training incorporates adversarial examples in the training stage to improve the robustness of DNNs\cite{goodfellow2015explaining,huang2015learning}. Unfortunately, almost all countermeasures, including adversarial training, are shown to be only effective to certain attack methods. They would likely not be defensive against some strong or unseen attacks\cite{wei2023self,wang2024defense}.

In this paper, the adversarial examples are crafted in the exactly opposite manner. The difference between the generated adversarial example and the original image is so large that people can hardly classify the adversarial example. However, the DNN still identifies the adversarial example as the same category as the original image.

Here comes the question of how the new type of adversarial example implements an attack to DNNs.  
If we consider the non-targeted attack as missing detection in object detection task, the proposed new type of adversarial example can be considered as a false alarm\cite{terzi2019directional}.  
In the former case, an attacker benefits from evading detection, while he profits from a fake target in the latter case\cite{zhao2023protecting,zhai2025adversarial}.  
For example, the new type of adversarial example can be used to attack identity authentication systems (e.g., face recognition systems) where it is passed off as an authorized user\cite{bensaid2025deep,qin2023adversarial}.  
Another potential application direction would be encryption. The new type of adversarial example can be utilized to hide image information since it almost looks like meaningless noise images, and the covert information can be extracted with a specified DNN.

Besides practical application value, the adversarial examples reveal some counterintuitive characteristics, or intrinsic blind spots of DNNs.  
Existing adversarial examples lie in the vicinity of a data point, which suggests that the decision boundary learned by the DNN should be expanded to involve these exceptional points.  
On the contrary, the adversarial examples proposed in this paper are distributed far away from the data point, indicating the decision boundary should shrink to exclude these outliers.  
Our method to generate adversarial examples is prompted by seeking for the perturbation which minimizes the loss with a distance constraint.  
Different from before, the distance is large enough to guarantee that the generated adversarial example is considerably different from the original input.  
We linearize the loss function and perturb the input iteratively along the gradients to solve the constrained optimization problem.  
Thus we propose the negative iterative fast gradient sign method (negI-FGSM) with $L_\infty$ norm bound and the negative iterative fast gradient method (negI-FGM) with $L_2$ norm bound. Another two attack methods, negative momentum iterative fast gradient sign method (negMI-FGSM) and negative momentum iterative fast gradient method (negMI-FGM) are formed by integrating momentum into negI-FGSM and negI-FGM respectively.  
To evaluate the effectiveness of our methods, we conduct extensive experiments on different networks trained on the ILSVRC2012 dataset.  
These experiments show that the adversarial example produced by our approach is significantly distinguished from but still identified by the network as the same class as the original input.  
In summary, this paper makes the following contributions:

\begin{enumerate}
    \item We introduce a new type of adversarial example, which behaves exactly opposite to existing adversarial examples and is hard to defend.
    \item We propose iterative gradient-based methods-negI-FGSM and negI-FGM, and momentum methods-negMI-FGSM and negMI-FGM to generate the new type of adversarial example, which perturb the input in the negative gradient or momentum direction.
    \item Our work shows that adversarial examples not only lie in the vicinity of a data point, but also are distributed far away from the data point where the learned decision boundary should contract.
\end{enumerate}

The rest of this paper is organized as follows: The background knowledge about adversarial attack is provided in Section II. We introduce the new type of adversarial example and propose the generating methods, including negI-FGSM, negI-FGM, negMI-FGSM, and negMI-FGM, in Section III. Section IV verifies the effectiveness of our methods through some experiments. Finally, we conclude the paper in Section V.

\section{Preliminaries}
In this section, we review the background and the related works on adversarial attack.
\subsection{Problem Formulation}
Given a DNN-based classifier $f(\textit{\textbf{X}})$ : $\textit{\textbf{X}} \in \mathcal{X} \to y \in \mathcal{Y}$ where $\textit{\textbf{X}}$ denotes an input image and $y$ is the classification result for $\textit{\textbf{X}}$.
The adversary aims to find an adversarial example $\textit{\textbf{X}}^{adv}$ which is misclassified by the DNN under an $\epsilon$-constraint, i.e., $\left\| \textit{\textbf{X}}^{adv} - \textit{\textbf{X}} \right\|_p  \le \epsilon$, where $p$ represents $L_p$ norm and could be chosen from 0, 1, 2, $\infty$.
$\epsilon$ is usually set sufficiently small to ensure that the perturbation is imperceptible.
Existing adversarial examples can be categorized into either untargeted or targeted ones.
For an input image $\textit{\textbf{X}}$ with ground-truth label $y_{true}$, suppose it is correctly classified by the DNN, that is, $f(\textit{\textbf{X}}) = y_{true}$. An untargeted adversarial example $\textit{\textbf{X}}^{adv}$ crafted from $\textit{\textbf{X}}$ misleads the classifier as $f(\textit{\textbf{X}}^{adv}) \ne y_{true}$, while
a targeted adversarial example fools the classifier to output a specific label $y^*$ such that $f(\textit{\textbf{X}}^{adv}) = y^*$, where $y^* \ne y_{true}$. We introduce the untargeted adversarial attacks here, and the targeted version can be easily derived.

Let $J(\textit{\textbf{X}}, y)$ denote the loss function, for example the cross-entropy loss in most cases.
An adversarial example can be found by maximizing $J(\textit{\textbf{X}}, y)$ under the $\epsilon$-constraint.
The adversarial attack is formulated as
\begin{equation}\label{eq.1}
    \mathop {\arg \max }\limits_{\textit{\textbf{X}}^{adv}} J(\textit{\textbf{X}}^{adv},y_{true} ) \quad {\rm{s}}{\rm{.t}}{\rm{.}} \quad \left\| \textit{\textbf{X}}^{adv} - \textit{\textbf{X}} \right\|_p  \le \epsilon.
\end{equation}
The above formulation renders $\textit{\textbf{X}}^{adv}$ most discriminative to the true class by the classifier.
\subsection{Attack Methods}
Methods that can solve the constrained optimization problem in (\ref{eq.1}) form the attack methods as below.

\textbf{One-step methods} perturb the input image in the gradient direction of $J(\textit{\textbf{X}}, y)$ where $J(\textit{\textbf{X}}, y)$ grows fastest. If it is optimized under the $L_\infty$ norm constraint, adversarial examples are generated as
\begin{equation}\label{eq.2}
    \textit{\textbf{X}}^{adv}  = \textit{\textbf{X}} + \epsilon \cdot {\rm{sign}}(\nabla _{\textit{\textbf{X}}} J(\textit{\textbf{X}},y_{true} )),
\end{equation}
where $\nabla _{\textit{\textbf{X}}}J(\textit{\textbf{X}},y_{true} )$ is the gradient of $J(\textit{\textbf{X}}, y_{true})$ w.r.t. $\textit{\textbf{X}}$. This method is called FGSM\cite{goodfellow2015explaining}. An adversarial example generated with FGSM can differ from the original image by at most $\epsilon$ at any pixel location.
The fast gradient method (FGM) generalizes FGSM to satisfy the $L_2$ norm bound $\left\| \textit{\textbf{X}}^{adv} - \textit{\textbf{X}} \right\|_2  \le \epsilon$ as
\begin{equation}\label{eq.3}
    \textit{\textbf{X}}^{adv}  = \textit{\textbf{X}} + \epsilon \cdot \frac{{\nabla _\textit{\textbf{X}} J(\textit{\textbf{X}},y_{true} )}}{{\left\| {\nabla _\textit{\textbf{X}} J(\textit{\textbf{X}},y_{true} )} \right\|_2 }}.
\end{equation}

\textbf{Iterative methods}\cite{kurakin2018adversarial} iteratively carry out the accumulation along the direction of gradient as in (\ref{eq.2}) and (\ref{eq.3}) with small step size.
For example, the iterative version of FGSM (I-FGSM) can be depicted as:
\begin{equation}\label{eq.4}
    {\textit{\textbf{X}}}_0^{adv}  = {\textit{\textbf{X}}}, \quad {\textit{\textbf{X}}}_{n + 1}^{adv}  = {\textit{\textbf{X}}}_n^{adv}  + \alpha \cdot{\rm{sign}}(\nabla _{\textit{\textbf{X}}} J({\textit{\textbf{X}}}_n^{adv} ,y_{true} )),
\end{equation}
where the step size $\alpha$ can be simply set as $\epsilon/N$ with $N$ being the maximum number of iteration to meet the $L_\infty$ bound. 
Alternatively, one can clip the intermediate results per pixel in each iteration into the $\epsilon$-neighbourhood of $\textit{\textbf{X}}$:
\begin{equation}\label{eq.5}
    {\textit{\textbf{X}}}_{n + 1}^{adv}  = 
    Clip_{\textit{\textbf{X}},\epsilon}
    \left\{ {{\textit{\textbf{X}}}_n^{adv}  + \alpha \cdot{\rm{sign}}(\nabla _{\textit{\textbf{X}}} J({\textit{\textbf{X}}}_n^{adv} ,y_{true} ))} \right\}.
\end{equation}
For adversarial attack methods, there is usually a trade-off between the attack ability and the transferability.
It has been proved that iterative methods exhibit superior attack effect in the white-box manner to one-step methods
at the cost of worse transferability\cite{kurakin2017adversarial,kurakin2018adversarial,tramer2018ensemble}.

\textbf{Optimization-based methods}\cite{szegedy2014intriguing} convert the constrained optimization problem in (\ref{eq.1}) to an unconstrained one in a way similar to the Lagrange multiplier method as\cite{carlini2017towards}
\begin{equation}\label{eq.6}
    \mathop {\arg \min }\limits_{\textit{\textbf{X}}^{adv}} \lambda  \cdot  \left\| \textit{\textbf{X}}^{adv} - \textit{\textbf{X}} \right\|_p - J(\textit{\textbf{X}}^{adv},y_{true} ). 
\end{equation}
Box-constrained L-BFGS can be employed to solve this problem\cite{szegedy2014intriguing}.
Optimization-based methods jointly optimize the loss function and the distance between the adversarial example and original image. Then the distance constraint changes into a soft constraint, i.e., the $L_p$ distance is not guaranteed to be smaller than the required value.
Since L-BFGS is a derivative-based iterative algorithm, optimization-based methods also have poor transferability just like iterative methods.
\section{New Type of Adversarial Example and Its Generation}
This section introduces the new type of adversarial example and presents the generating methods of it.
\subsection{New Type of Adversarial Example}
The new type of adversarial example behaves in the completely opposite way to the existing adversarial example. Specifically, the new type of adversarial example is crafted to be significantly different from the original image.
Therefore, the new type of adversarial example is generated under the $L_p$ norm constraint
$\left\| \textit{\textbf{X}}^{adv} - \textit{\textbf{X}} \right\|_p  \ge \delta $, where 
$\delta $ is set large enough to guarantee that the difference between the adversarial example and the original image is significant. 
However, the DNN still identifies the adversarial example as the same class as the original image such that $f(\textit{\textbf{X}}^{adv}) = y_{true}$, supposing the original image is correctly classified.
To this end, the loss function $J(\textit{\textbf{X}}, y)$ should be minimized subject to the $\delta$-constraint, i.e.,
\begin{equation}\label{eq.7}
    \mathop {\arg \min }\limits_{\textit{\textbf{X}}^{adv}} J(\textit{\textbf{X}}^{adv},y_{true} ) \quad {\rm{s}}{\rm{.t}}{\rm{.}} \quad \left\| \textit{\textbf{X}}^{adv} - \textit{\textbf{X}} \right\|_p  \ge \delta.
\end{equation}
The adversarial example found according to (\ref{eq.7}) looks obviously different from the original image but is likely to be identified as the same class by the DNN.
\subsection{Adversarial Example Generation Methods}
Method for generating the new type of adversarial examples is also found by solving the constrained minimization problem in (\ref{eq.7}).
However, one-step methods are no longer competent to tackle the problem since one-step linear approximation in the large $\delta$-neighbourhood is infeasible.
Thus, we modify the iterative methods to form generation method of the new type of adversarial examples.
Specifically, we propose negI-FGSM, a variant of I-FGSM, which perturbs the input along the negative gradient direction:
\begin{equation}\label{eq.8}
    {\textit{\textbf{X}}}_0^{adv}  = {\textit{\textbf{X}}}, \quad {\textit{\textbf{X}}}_{n + 1}^{adv}  = {\textit{\textbf{X}}}_n^{adv}  - \alpha \cdot{\rm{sign}}(\nabla _{\textit{\textbf{X}}} J({\textit{\textbf{X}}}_n^{adv} ,y_{true} )).
\end{equation}
We can set the maximum number of iteration $N$ or compare the $L_\infty$ norm $\left\| \textit{\textbf{X}}_n^{adv} - \textit{\textbf{X}} \right\|_\infty$ with $\delta$ to determine the termination of iteration.
The step size $\alpha$ can be set as $\delta/N$ or any small value to guarantee the justification of linearization.
When the maximum number of iteration is set as $N$ and $\alpha$ is set as $\delta/N$, the $L_\infty$ distance $\left\| \textit{\textbf{X}}_N^{adv} - \textit{\textbf{X}} \right\|_\infty$ is not ensured to be larger than $\delta$. However, it does not matter as long as $\delta$ is set sufficiently large to make sure that the generated image is different substantially from the original one.
Besides, one can set $\left\| \textit{\textbf{X}}_n^{adv} - \textit{\textbf{X}} \right\|_\infty  \ge \delta $ as the condition for iteration termination to meet the $L_p$ norm constraint strictly.
To find an adversarial example under the constraint of $L_2$ norm bound $\left\| \textit{\textbf{X}}^{adv} - \textit{\textbf{X}} \right\|_2  \ge \delta $, negI-FGSM can be extended to negative iterative fast gradient method (negI-FGM) as
\begin{equation}\label{eq.9}
    {\textit{\textbf{X}}}_{n + 1}^{adv}  = {\textit{\textbf{X}}}_n^{adv} - \alpha \cdot \frac{{\nabla _{\textit{\textbf{X}}} J({\textit{\textbf{X}}}_n^{adv} ,y_{true} )}}{{\left\| {\nabla _{\textit{\textbf{X}}} J({\textit{\textbf{X}}}_n^{adv} ,y_{true} )} \right\|_2 }}.
\end{equation}
However, negI-FGSM (or negI-FGM) greedily updates the input and is more likely to fall into local minimum. To escape from local optimum, a momentum term\cite{polyak1964some} is integrated into negI-FGSM, forming a new attack method named negative momentum iterative fast gradient sign method (negMI-FGSM). The update procedure of negMI-FGSM is formulated as:
\begin{align}
    {{\boldsymbol{g}}_{n + 1}} &= \mu  \cdot {{\boldsymbol{g}}_n} + \frac{{{\nabla _{\textit{\textbf{X}}}}J({\textit{\textbf{X}}}_n^{adv},{y_{true}})}}{{{{\left\| {{\nabla _{\textit{\textbf{X}}}}J({\textit{\textbf{X}}}_n^{adv},{y_{true}})} \right\|}_1}}}, \label{eq.10}\\
    {\textit{\textbf{X}}}_{n + 1}^{adv} &= {\textit{\textbf{X}}}_n^{adv} - \alpha  \cdot {\rm{sign}}({{\boldsymbol{g}}_{n + 1}}),\label{eq.11}
\end{align}
where $\mu$ in (\ref{eq.10}) is the decay factor and negMI-FGSM degenerates to negI-FGSM when $\mu=0$. ${\boldsymbol{g}}_{n+1}$ accumulates the normalized gradients of the first $n+1$ iterations with ${{\boldsymbol{g}}_{0}}=0$.
The accumulation helps to accelerate gradient descent algorithms and barrel through local optimum, small humps and narrow valleys, which will better guarantee the attack effect\cite{duch1998optimization}.
It is worth noting that another advantage of the momentum method is better stability in the iteration process of stochastic gradient descent algorithm\cite{sutskever2013importance,qian1999momentum}.
Then the intermediate result at the $n$-th iteration ${\textit{\textbf{X}}}_n^{adv}$ is updated by adding perturbation in the negative direction of the sign of ${\boldsymbol{g}}_{n+1}$ with a step size $\alpha$ in (\ref{eq.11}). 
By substituting the current gradient with the momentum term ${\boldsymbol{g}}_{n+1}$, any iterative method can be generalized to its momentum variant.
The momentum variant of negI-FGM, named negMI-FGM can be expressed as
\begin{equation}\label{eq.12}
    {\textit{\textbf{X}}}_{n + 1}^{adv} = {\textit{\textbf{X}}}_n^{adv} - \alpha \cdot  \frac{{{\boldsymbol{g}}_{n + 1}}}{{\left\| {{\boldsymbol{g}}_{n + 1}} \right\|_2 }}.
\end{equation}

\section{Experiments}
We perform a series of comprehensive experiments to evaluate the attack effect of the proposed methods under different hyperparameters.
\subsection{Setup}
We investigate four models: Inception v3 (Inc-v3)\cite{szegedy2016rethinking}, Inception v4 (Inc-v4), Inception-Resnet v2 (IncRes-v2)\cite{szegedy2017inception}, and Resnet v2-152 (Res-152)\cite{he2016identity}, which are all normally trained.

It seems meaningless to evaluate the attack effect if the models are unable to correctly classify the original image. Thus, we randomly select 1000 images belonging to the 1000 categories from the ILSVRC 2012 validation set\cite{russakovsky2015imagenet}, all of which are correctly classified by the four models.

The vanilla iterative methods, negI-FGSM and negI-FGM have two hyperparameters—the size of perturbation and the number of iterations, while the momentum-based iterative methods, negMI-FGSM and negMI-FGM have an extra hyperparameter—the decay factor. We conduct the following ablation experiments to evaluate the success rates of adversarial attacks against the four models under different hyperparameter settings, from which we can find the impact of these hyperparameters on the attack effect of the proposed methods. The higher correctly classification rates, the better the attack effect.

\subsection{Size of perturbation}
We study the effects of different perturbation sizes on the success rates of attacks. We generate adversarial examples using the Inc-v3 model with four attack methods: negI-FGSM, negI-FGM, negMI-FGSM, and negMI-FGM. The perturbation sizes range from 8 to 80 in the pixel value [0, 255]. The number of iterations is 250, and the decay factor is 0.8.

\begin{figure*}[!t]
\centering
\includegraphics[width=4.8in]{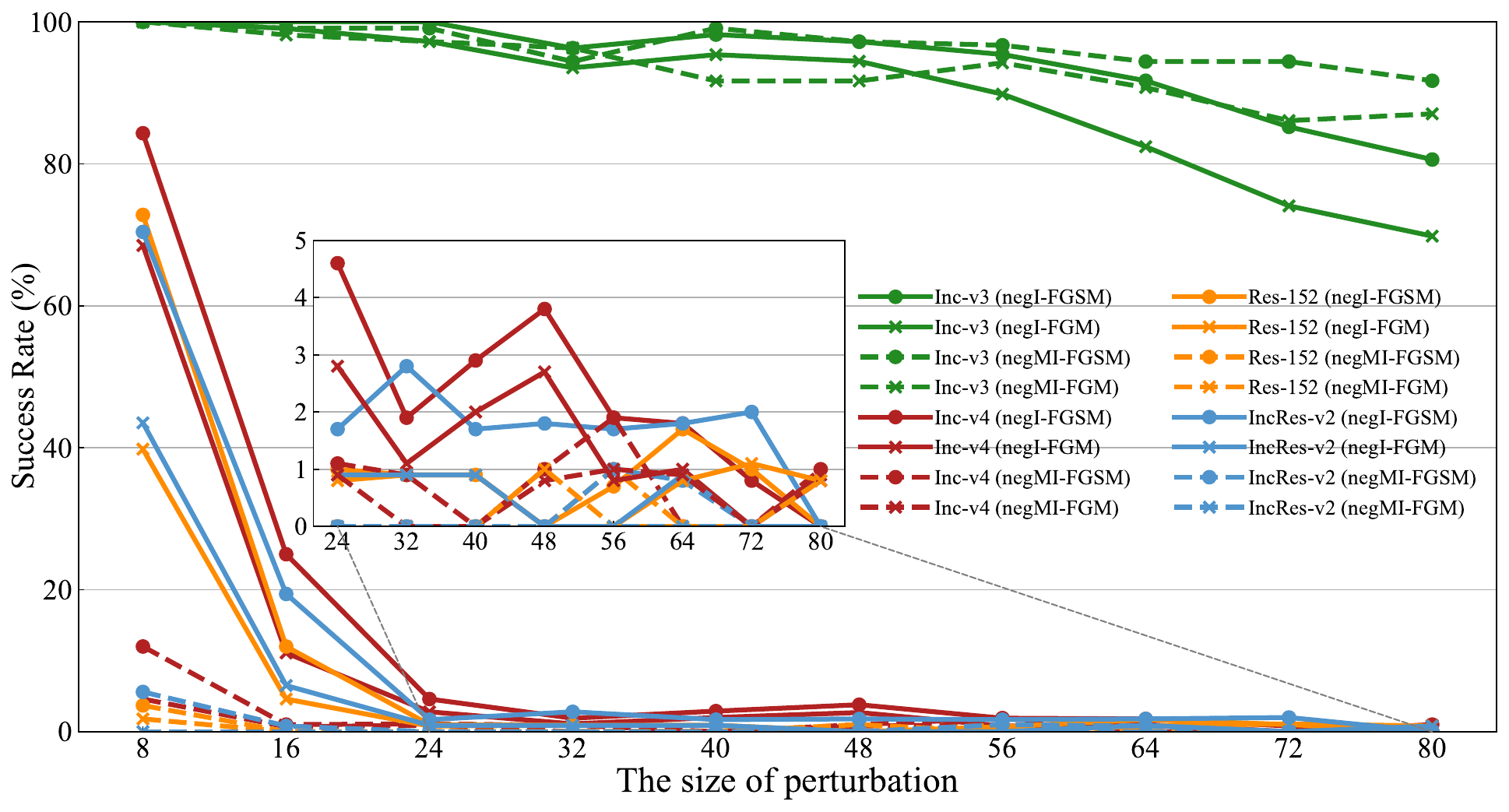}
\caption{Success rates of adversarial examples generated for the Inc-v3 model against different models: white-box model for Inc-v3, and black-box models for Inc-v4, IncRes-v2,and Res-152. We compare the results of four methods with different sizes of perturbation.}
\label{fig_1}
\end{figure*}

Fig.\ref{fig_1} illustrates the success rates of adversarial attacks against the white-box model Inc-v3 and three black-box models--Inc-v4, IncRes-v2, and Res-152. In a white-box attack setting, the success rates of all four attack methods decrease as the perturbation sizes increase. When the perturbation size is below 16, all four attack methods achieve success rates of nearly $100\%$. However, when the perturbation increases to 80, the negMI-FGSM method achieves a success rate of approximately $91\%$, the negMI-FGM method achieves around $87\%$, the negI-FGSM method achieves approximately $81\%$, and the negI-FGM method achieves an attack success rate of approximately $65\%$. This decrease in success rates is attributed to the significant shifts in the positions of adversarial examples within the feature space, making it easier for them to cross the model's decision boundaries and lead to misclassification. In black-box attack settings, the model shows high recognition accuracy for adversarial examples when the perturbation size is small. However, as the perturbation size increases, the model's recognition accuracy for these adversarial examples drops sharply below $5\%$. The phenomenon is mainly due to the significant differences in decision boundaries between different models. 

\begin{figure}[!t]
\centering
\includegraphics[width=3.49in]{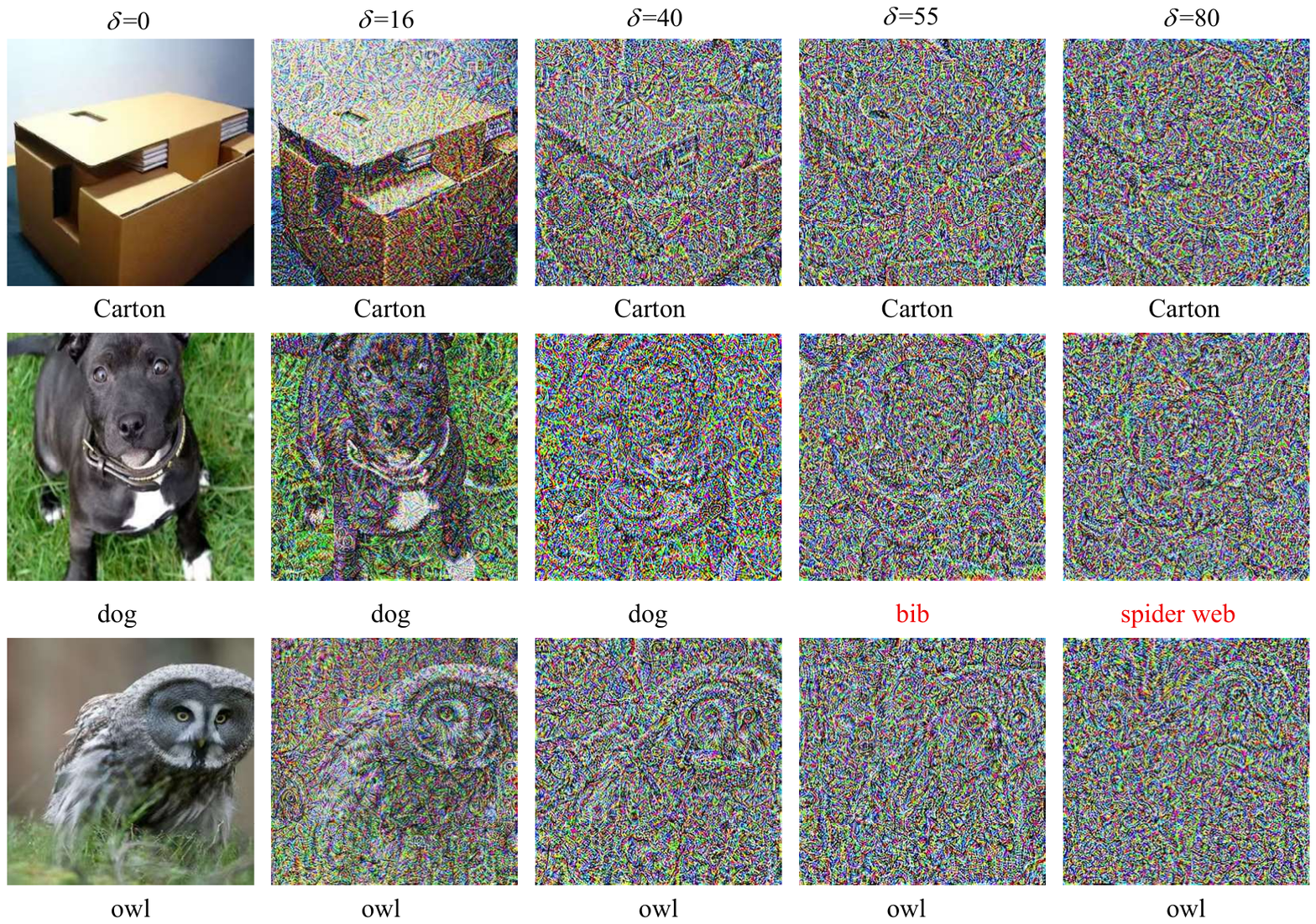}
\caption{Comparison of adversarial examples under different perturbations}
\label{fig_2}
\end{figure}

Fig.\ref{fig_2} visualizes adversarial examples generated for Inc-v3 using negI-FGSM with different perturbation sizes. When the perturbation size reaches 16, the details of the image begin to degrade, but the main contour features are still visually recognizable. When the perturbation is set to 40, the image becomes sufficiently blurred and exhibits significant visual differences from the original image. The adversarial example remains within the decision boundary of the specified DNN, which can still be classified accurately. However, when the perturbation size reaches 55 or 80, some adversarial examples cross the decision boundary of the DNN in the feature space, leading to the adversarial examples being misclassified by the model. The figure illustrates that when the perturbation is set to 55, the Inc-v3 model incorrectly classifies a dog-class adversarial example as a "bib." Additionally, when the perturbation is increased to 80, the Inc-v3 model misclassifies the same dog-class adversarial example as a "spider web."

\begin{figure}[!t]
\centering
\includegraphics[width=3.49in]{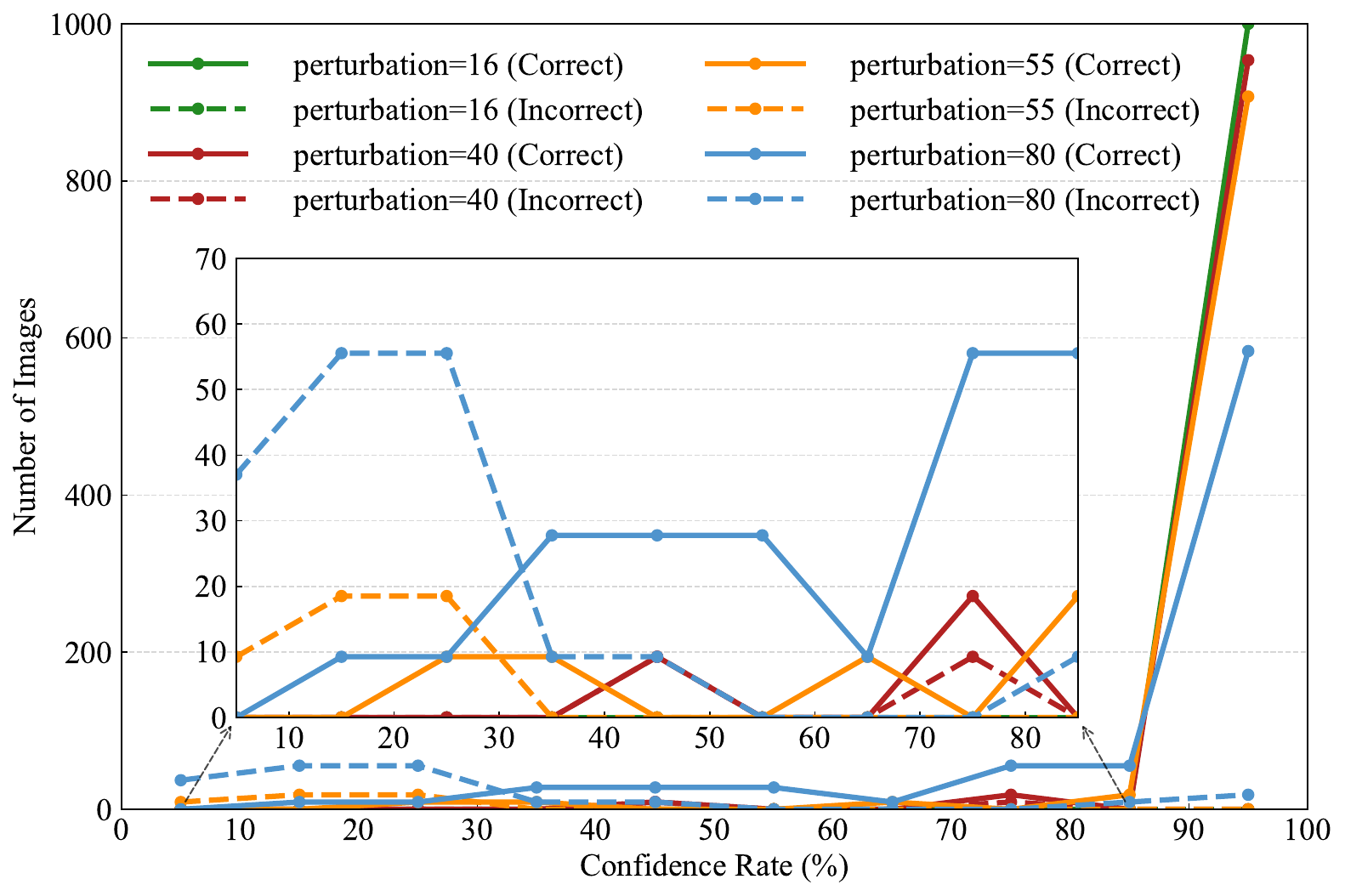}
\caption{The figure shows the trend of the number of correctly and incorrectly classified adversarial examples with confidence under different perturbations, where the adversarial examples are  generated for Inc-v3 using negI-FGSM in a white-box setting.}
\label{fig_3}
\end{figure}
Fig.\ref{fig_3} presents trends in the number of correctly and incorrectly classified adversarial examples across varying confidence levels under different perturbations, with the adversarial examples generated for Inc-v3 using the negI-FGSM method. In the low confidence interval, the number of correctly classified adversarial examples is low; while in the high confidence interval, the number of correctly classified adversarial examples increases significantly, and the model is able to correctly classify almost all adversarial examples when the amount of perturbation is small. For example, when the perturbation value is 16, the model correctly classifies 1000 adversarial examples in the high confidence interval; however, as the perturbation value increases to 80, the number of correctly classified examples in the high confidence interval decreases to about 800. This phenomenon suggests that even though the adversarial examples experience a large amount of perturbation, the model is still able to maintain a high classification accuracy when classifying these examples.

\subsection{Number of Iterations}
We evaluate the effects of the number of iterations on success rates. We generate adversarial examples using the Inc-v3 model with four attack methods: negI-FGSM, negI-FGM, negMI-FGSM, and negMI-FGM. The number of iterations ranges from 50 to 400 in steps of 50, with the perturbation size fixed at 80 and the decay factor set to 0.8.

\begin{figure*}[!t]
\centering
\includegraphics[width=4.8in]{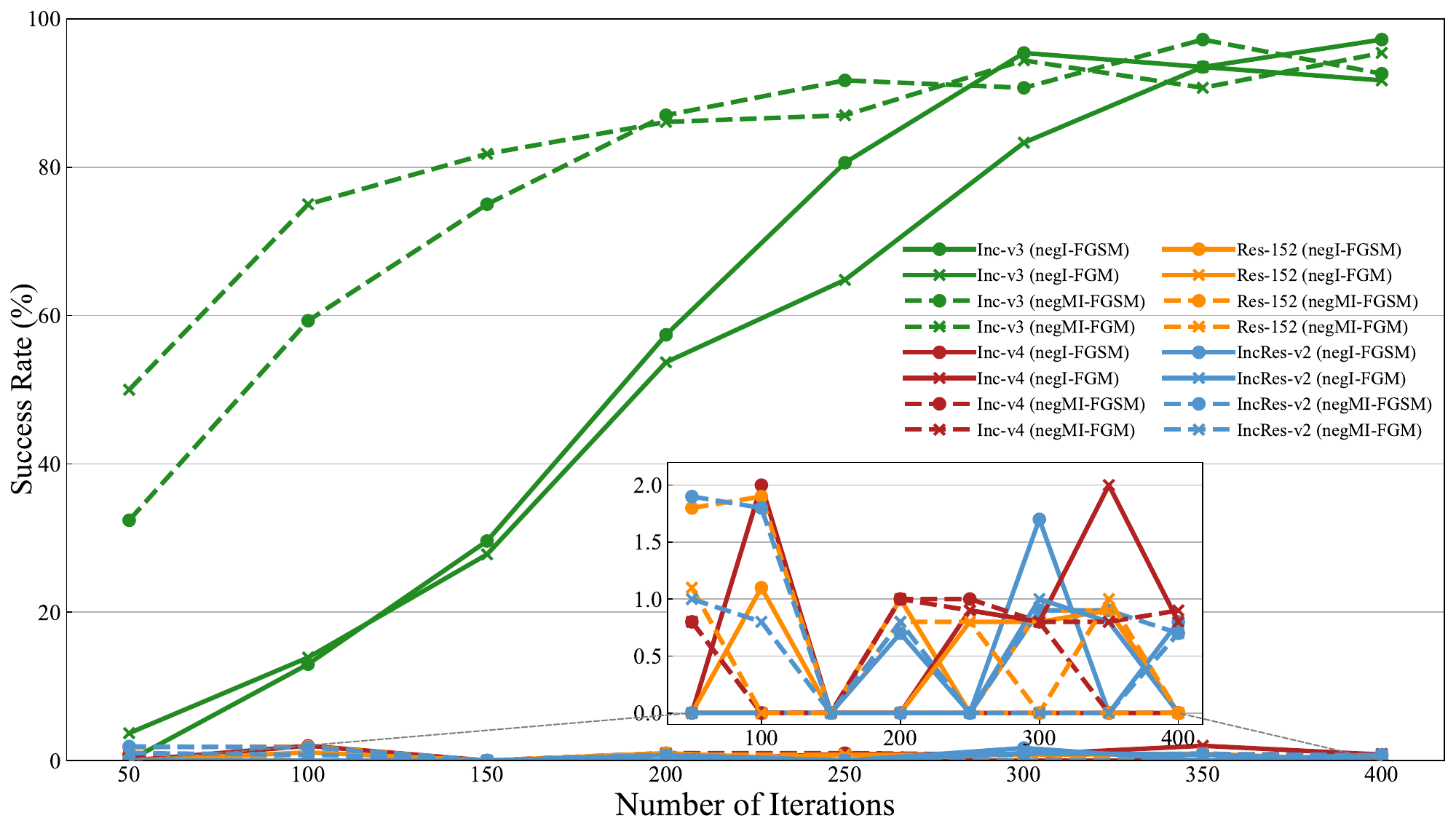}
\caption{The success rates of adversarial examples created for the Inc-v3 model are tested on different models: the white-box model (Inc-v3) and the black-box models (Inc-v4, IncRes-v2, and Res-152). We compare the results of four methods with different numbers of iterations.}
\label{fig_4}
\end{figure*}

\begin{figure}[!t]
\centering
\includegraphics[width=3.49in]{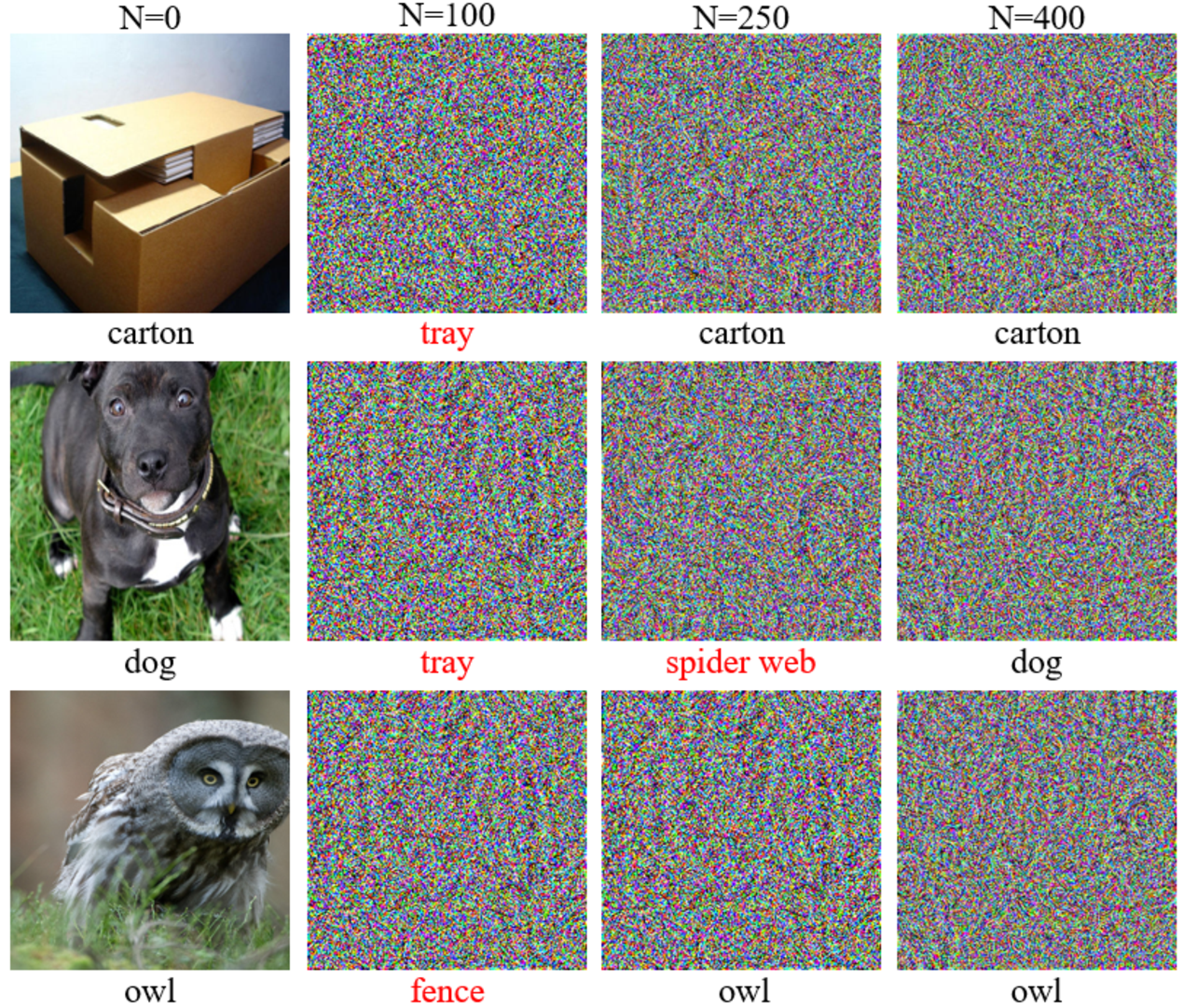}
\caption{Comparison of adversarial examples under different iterations}
\label{fig_5}
\end{figure}

Fig.\ref{fig_4} illustrates the success rates of adversarial attacks against the white-box model Inc-v3 and three black-box models—Inc-v4, IncRes-v2, and Res-152. In a white-box attack setting, the success rates of all four attack methods rise as the number of iterations increases. When the number of iterations increases to 400, the negMI-FGSM method achieves an attack success rate of approximately $92\%$, the negI-FGSM method achieves approximately $97\%$, the negI-FGM method achieves approximately $92\%$, and the negMI-FGM method achieves approximately $95\%$. When the number of iterations is low, the attack methods assume that the decision boundary around the data point is linear, making it difficult to accurately capture the complex nonlinear behaviour in DNNs. As the number of iterations increases, the attack methods gradually approach the model's decision boundary by continuously adjusting the gradient direction. In the black-box setting, the success rates of all four attack methods remain low as the number of iterations increases. For example, when the number of iterations is set to 50, the adversarial examples generated for the Inc-v3 model by negI-FGM are completely misclassified by the Inc-v4 model. When the number of iterations increases to 400, the recognition accuracy of the black-box model on these adversarial examples is still as low as $2\%$, which indicates that the new type of adversarial examples can only be correctly recognized by a specified DNN.

Fig.\ref{fig_5} illustrates adversarial examples generated for Inc-v3 using negI-FGSM with different numbers of iterations and a fixed perturbation size of 80. When the number of iterations exceeds 100, the images become completely blurred, making it visually impossible to extract any useful information. However, the attack method relies on the assumption of linearity in the decision boundary and struggles to optimize its attack direction effectively with a low number of iterations. When the number of iterations reaches 100, the Inc-v3 model misclassifies the original "carton" and "dog" adversarial examples as "tray," and the "owl" adversarial example is misclassified as "fence." When the number of iterations increases to 250, the adversarial example of "dog" is misclassified as "spider web" by the Inc-v3 model. When the number of iterations reaches 400, the Inc-v3 model correctly classifies several adversarial examples that it misclassified at lower iterations. 

\begin{figure}[!t]
\centering
\includegraphics[width=3.49in]{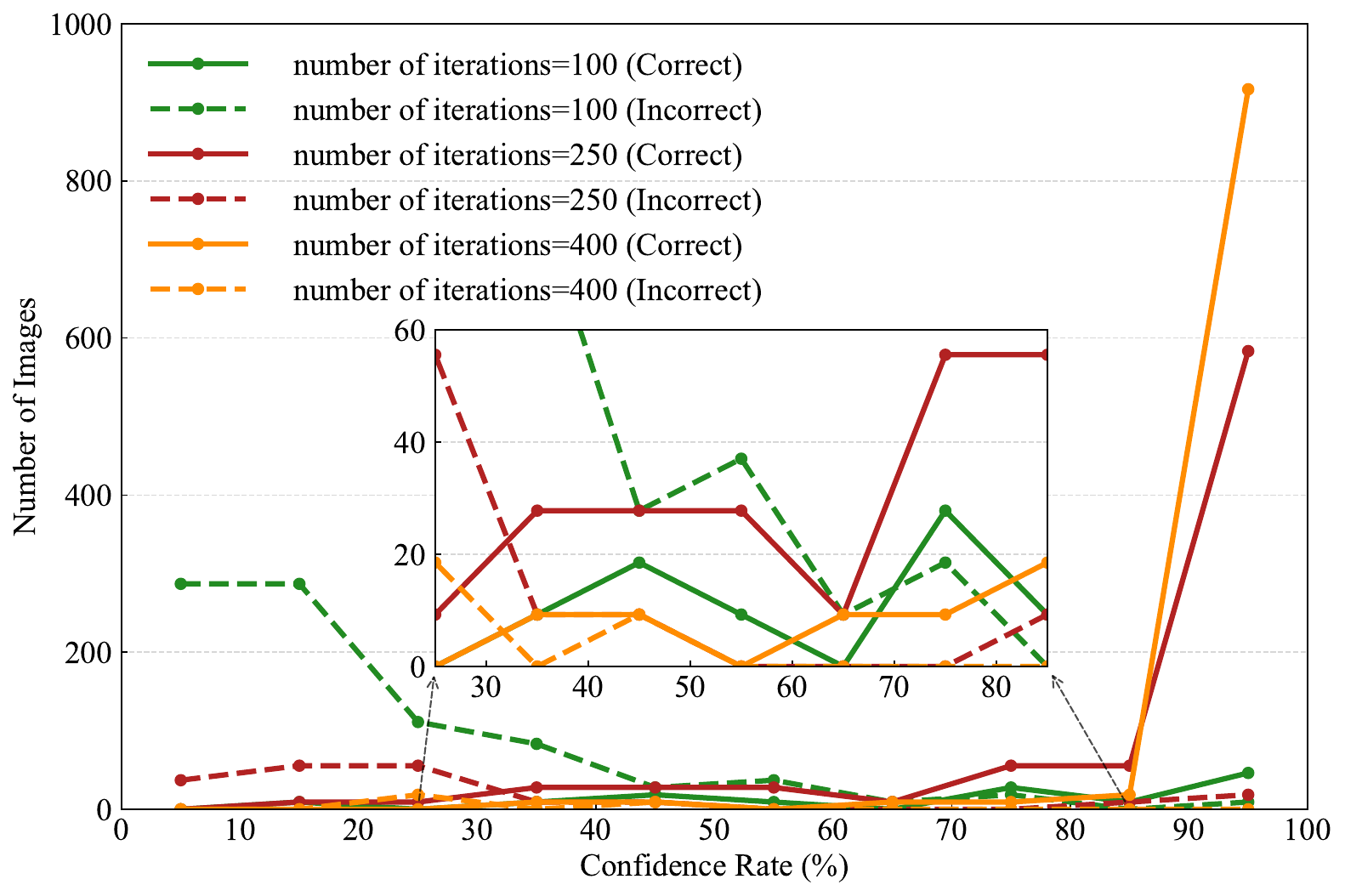}
\caption{The figure shows the trend of the number of correctly and incorrectly classified adversarial examples with confidence under different iterations, where the adversarial examples are generated for Inc-v3 using negI-FGSM in a white-box setting.}
\label{fig_6}
\end{figure}

Fig.\ref{fig_6} illustrates the trend of the number of correctly classified and misclassified adversarial examples as the confidence level changes in different iteration numbers, and the adversarial examples are created for Inc-v3 using negI-FGSM in a white-box setting. In the low-confidence region, the number of misclassification errors made by the model for the adversarial examples is significantly higher than the number of correctly classified adversarial examples, while in the high-confidence region, the number of correctly classified adversarial examples is significantly higher than the number of incorrectly classified adversarial examples when the number of iterations is 250 and 400. It is worth noting that when the number of iterations is 100, the correct classification rate of the adversarial examples is still low despite being in the higher confidence region, a phenomenon that suggests that as the number of iterations increases, the attacking method, by constantly adjusting the direction of the gradient, allows the specified model to correctly identify the adversarial examples even though it is visually difficult to extract useful information from the examples.

\subsection{Decay factor $\mu$}
We explore the impact of the decay factor on the success rates of adversarial examples. We generate adversarial examples for the Inc-v3 model using momentum-based methods--negMI-FGSM and negMI-FGM with a perturbation size of 80, the number of iterations 250, and the decay factor $\mu$ ranging from 0.0 to 1.4 in steps of 0.2.

\begin{figure}[!t]
\centering
\includegraphics[width=3.49in]{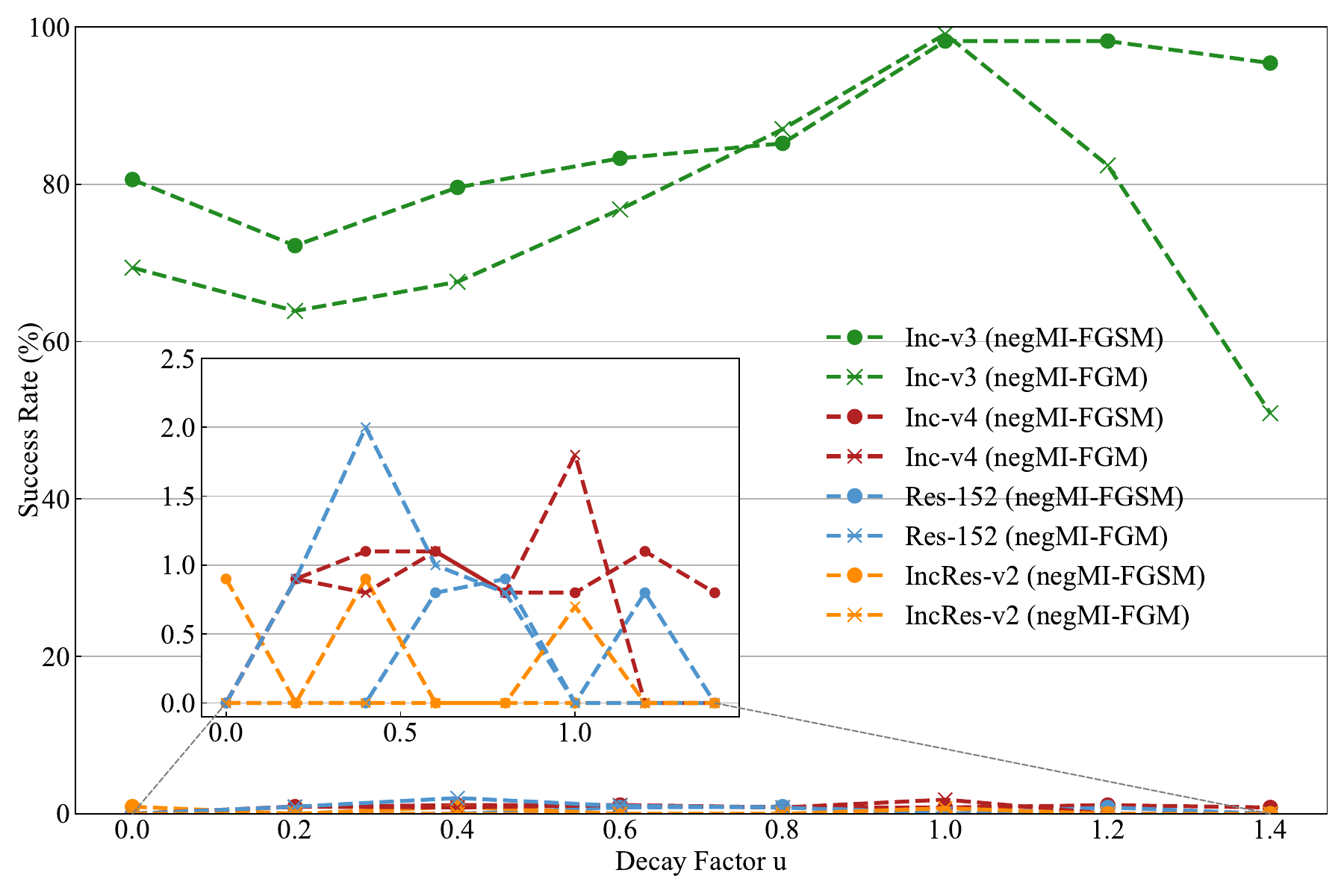}
\caption{Success rates of adversarial examples generated for the Inc-v3 model against different models: white-box method for Inc-v3, and black-box methods for Inc-v4, IncRes-v2, and Res-152, with \( u \) ranging from 0.0 to 1.4.}
\label{fig_7}
\end{figure}

Fig.\ref{fig_7} illustrates the success rates of adversarial attacks against the white-box model Inc-v3 and three black-box models—Inc-v4, IncRes-v2, and Res-152. In the white-box setting, the success rates of both attack methods increase as the decay factor approaches $1.0$ but begin to decrease when the decay factor exceeds $1.0$. When the decay factor is $0.0$, the attack success rate of negMI-FGM is $64\%$, and that of negMI-FGSM is $81\%$. When the decay factor is $1.0$, both attack methods achieve an attack success rate of $99\%$. However, when the decay factor increases to $1.4$, the attack success rate of negMI-FGSM decreases to $95\%$, while that of negMI-FGM drops to $51\%$. When the decay factor is $0.0$, the momentum method degenerates to the iterative method. When the decay factor is set to $1.0$, the momentum update is based on the accumulation of all previous gradients. This trend indicates that the introduction of momentum aids in smoothing the gradient information and helps to avoid local optima. However, if the decay factor becomes too large, excessive accumulation of historical gradients can obscure the useful information from current gradients, resulting in decreased success rates. In the black-box attack setting, the success rates of the models on adversarial examples remain below $2\%$ as the decay factor increases.

\begin{figure}[!t]
\centering
\includegraphics[width=3.49in]{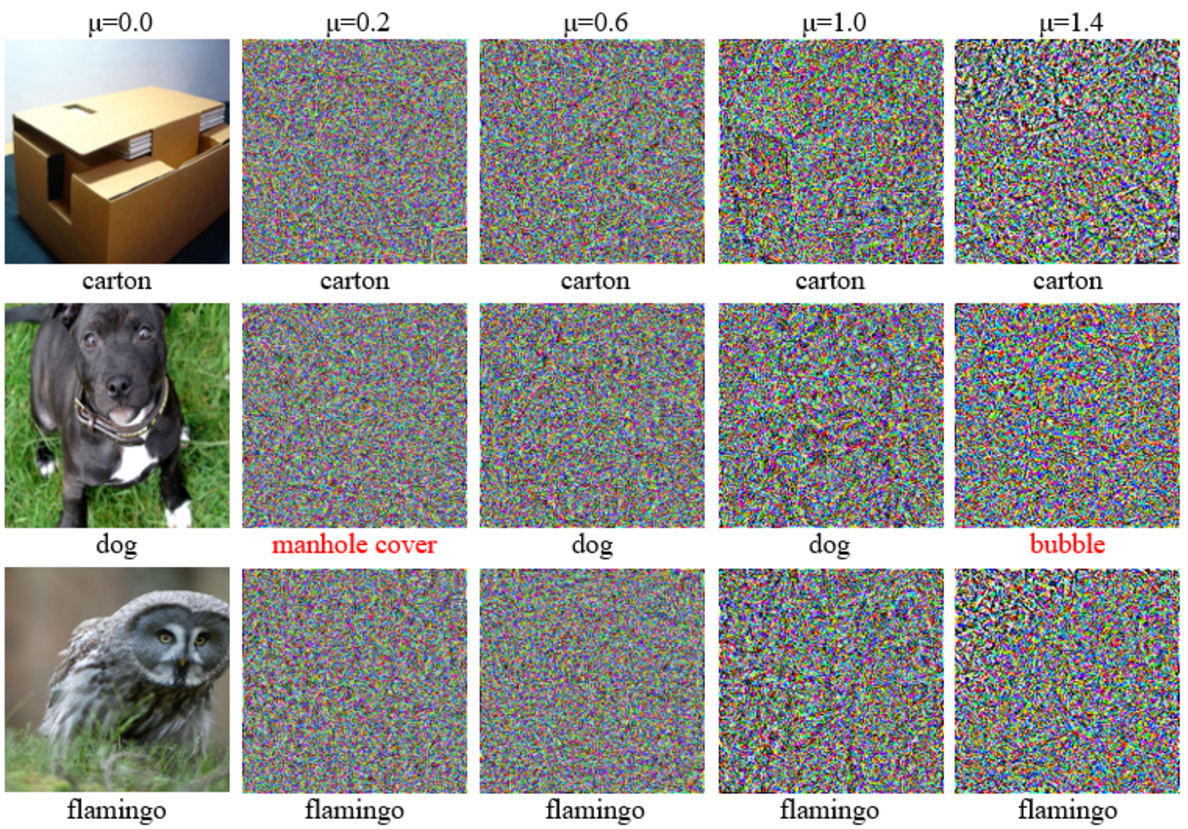}
\caption{Comparison of adversarial examples under different decay factor $\mu$}
\label{fig_8}
\end{figure} 

Fig.\ref {fig_8} illustrates adversarial examples generated for Inc-v3 using negMI-FGSM with decay factor $\mu$, a perturbation size fixed at 80, and the number of iterations set to 250. When the decay factor $\mu$ is small, the attack method becomes trapped in local minima during optimization, leading to misclassification by the model. For example, when $\mu=0.2$, the Inc-v3 model incorrectly classifies "dog" adversarial examples as "manhole cover." On the other hand, when the decay factor is excessively large, the model's accuracy on adversarial examples decreases due to the excessive accumulation of historical gradients, which obscures the current gradient information. For example, when the decay factor is $1.4$, the Inc-v3 model misclassifies adversarial examples originally labelled as "dog" as "bubble." 

\begin{figure}[!t]
\centering
\includegraphics[width=3.49in]{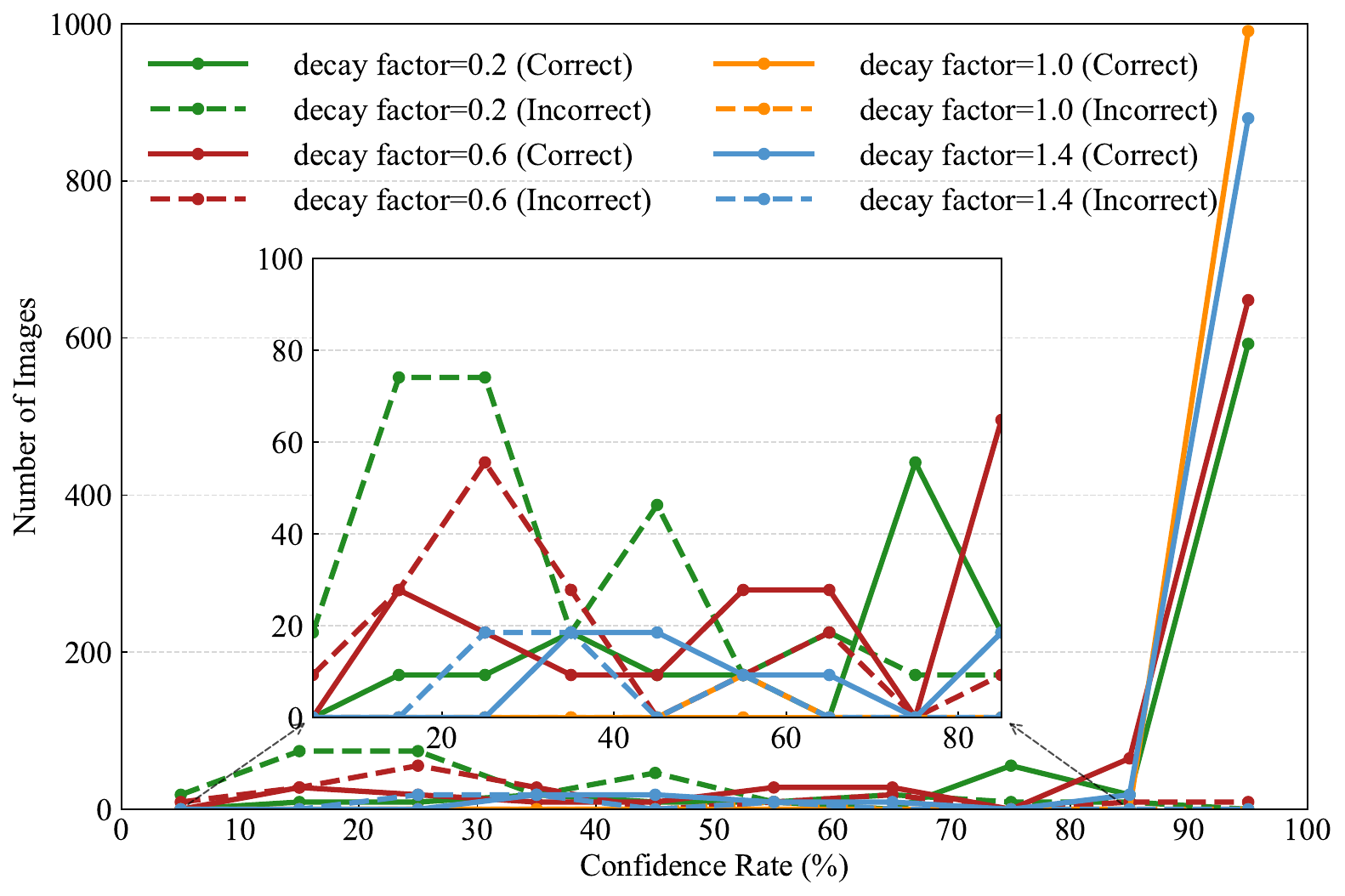}
\caption{The figure illustrates the trends of the quantities of correctly and incorrectly predicted examples as a function of confidence level for adversarial examples generated for Inc-v3 using negMI-FGSM in a white-box setting with different decay factors.}
\label{fig_9}
\end{figure}

\begin{figure}[!t]
\centering
\includegraphics[width=3.2in]{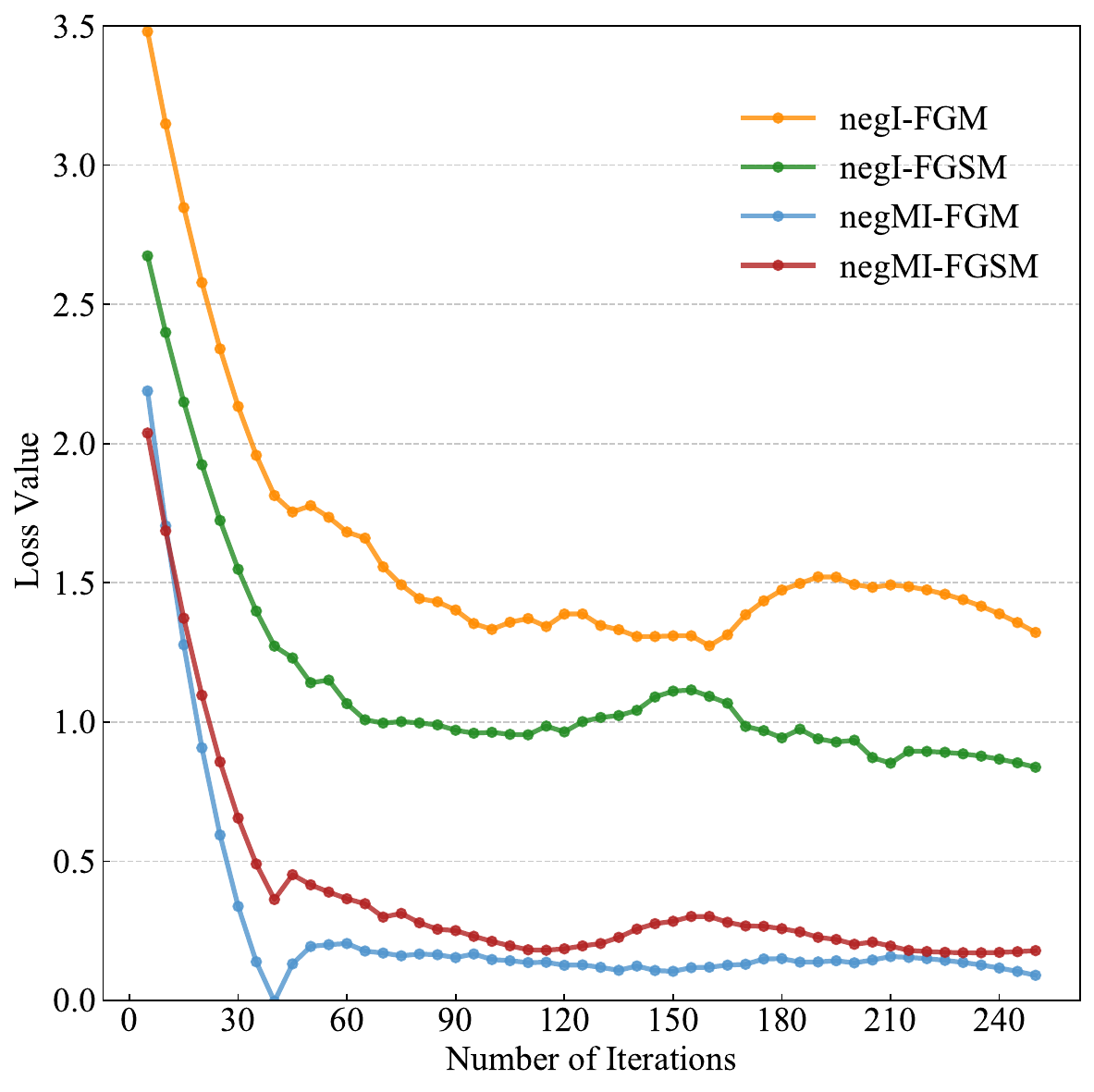}
\caption{The figure illustrates the variation trends of the loss values across different iterations. These trends are observed for adversarial examples generated for Inc-v3 using negI-FGM, negI-FGSM, negMI-FGM, and negMI-FGSM in a white-box setting.}
\label{fig_10}
\end{figure}
Fig.\ref {fig_9} depicts the trend of the number of positive and incorrectly predicted examples with the confidence level of the adversarial examples generated by the negMI-FGSM method under the white-box setting with different decay factors. With different decay factors considered, for the low-confidence interval, a small decay factor leads to a significantly higher quantity of misclassified adversarial examples compared to correctly classified ones. However, as the decay factor increases, in the high-confidence interval, the quantity of misclassified adversarial examples is much lower than that of correctly classified ones. Especially in the high-confidence interval, larger decay factors result in higher confidence levels of the adversarial examples. This indicates that selecting an appropriate decay factor can smooth the gradient information and prevent the model from converging to local optima.

In Fig.\ref{fig_10}, we present the trend of the values of the loss function as the number of iterations increases. It is evident that after incorporating momentum into the optimization process, the convergence performance of negMI-FGSM and negMI-FGM significantly improves compared to negI-FGSM and negI-FGM. This enhanced convergence behavior indicates that negMI-FGSM and negMI-FGM are more effective in handling the optimization problem, demonstrating their superior ability to navigate the complex loss landscape and reach a more optimal solution.

\subsection{Comparison of negI-FGSM, negI-FGM, negMI-FGSM and negMI-FGM}
Table \ref {table:1} presents the success rates of adversarial attacks against the white-box model Inc-v3 and three black-box models–Inc-v4, IncRes-v2, and Res-152. The adversarial examples are crafted using negI-FGSM, negI-FGM, negMI-FGSM, and negMI-FGM. The perturbation size is set to 80, the number of iterations is set to 250, and the decay factor is set to 0.8. A higher correctly classification rates of the model with adversarial examples as original images indicates a more effective attack strategy.

\begin{table}
\centering
\scriptsize 
\renewcommand{\arraystretch}{1.2}
\caption{}
\begin{tabular}{c|c|c|c|c|c}
\hline
\textbf{} & \textbf{Method} & \textbf{Inc-v3} & \textbf{Inc-v4} & \textbf{Res-152} & \textbf{IncRes-v2} \\
\hline
\hline
\multirow{4}{*}{Inc-v3} 
     & negI-FGSM  & 80.6\textsuperscript{*} & 0    & 0    & 0  \\
     & negI-FGM   & 64.8\textsuperscript{*} & 0.9 & 0.9    & 0.7 \\ 
     & negMI-FGSM & \textbf{91.7\textsuperscript{*}} & 0 & 0   & 0 \\
     & negMI-FGM  & 87.0\textsuperscript{*} & 0.7    & 0   & 0 \\
\hline
\multirow{4}{*}{Inc-v4}  
     & negI-FGSM  & 3.8  & 90.7\textsuperscript{*} & 0.8 & 3.6  \\
     & negI-FGM   & 1.0    & 81.5\textsuperscript{*} & 0   & 0 \\
     & NMI-FGSM & 0.9 & 92.6\textsuperscript{*} & 0.8   & 0 \\
     & negMI-FGM  & 0.7  & \textbf{94.4\textsuperscript{*}} & 0 & 0 \\
\hline    
\multirow{4}{*}{Res-152}
     & negI-FGSM  & 0    & 2.0 & 87.0\textsuperscript{*} & 0.9  \\
     & negI-FGM   & 1.0  & 0.8 & 79.6\textsuperscript{*} & 0.9  \\
     & negMI-FGSM & 0    & 0 & \textbf{88.9\textsuperscript{*}} & 0 \\
     & negMI-FGM  & 1.0  & 0.8 &80.6\textsuperscript{*} & 0 \\
\hline
\multirow{4}{*}{IncRes-v2}
     & negI-FGSM  & 0& 0.8 & 0 & 63.9\textsuperscript{*}  \\
     & negI-FGM   & 0.9 & 1.0 & 0 & 43.5\textsuperscript{*}  \\
     & negMI-FGSM & 0 & 1.7 & 0 & \textbf{78.1\textsuperscript{*}} \\
     & negMI-FGM  & 0 & 0 & 0 & 65.7\textsuperscript{*} \\
\hline                   
\end{tabular}
\begin{tabular}{l}
\toprule[0pt]
We evaluate the success rate (\%) of adversarial attacks against four models. The adv\\-ersarial examples are generated for Inc-v3, Inc-v4, IncRes-v2, and Res-152 using \\ negI-FGSM, negI-FGM, negMI-FGM, and negMI-FGSM.* denotes white-box attacks.
\end{tabular}

\label{table:1}
\end{table}
In the white-box attack settings, the Inc-v4 model exhibits superior classification capability for such new types of adversarial examples. Experiments show that when generating adversarial examples against Inc-v4 using the four attack methods, the success rates all exceed $80\%$, with the negMI-FGSM method achieving an attack success rate as high as $92.6\%$. In contrast, the IncRes-v2 model exhibits lower classification performance. Specifically, when adversarial examples are crafted against IncRes-v2 using the negI-FGM method, the model's correct classification rate is only $43.5\%$. In the black-box setting, the success rates of all four attack methods remain below $4\%$. For example, when adversarial examples are generated using the negI-FGSM method for Inc-v3, the Inc-v4 model achieves a correct classification rate of $3.8\%$.

Notably, the success rate in a white-box setting is much higher than in black-box settings, which indicates that the novel adversarial examples can obscure image information, rendering it recognizable solely by the specified DNN. On the other hand, the introduction of momentum better guarantees the attack effect. For example, in white-box settings, methods that use momentum, like negMI-FGSM and negMI-FGM, are more successful with different models than methods that rely on iterative gradients, such as negI-FGSM and negI-FGM. In black-box settings, momentum-based methods exhibit lower success rates than iterative gradient-based methods.


\begin{table*}[!t]
\caption{}
\label{table:2}
    \centering
    \scriptsize
    \renewcommand{\arraystretch}{1.2}
    \begin{tabular}{l|c|c|c|c|c}
    \hline
    & Method& negI-FGSM & negMI-FGSM & negI-FGM & negMI-FGM \\
    \hline
    \hline
    
    \multirow{3}{*}{Inc-v3}
    & SSIM $\uparrow$        & 0.0065 & 0.0049 & 0.0069 & 0.0045 \\
    & PSNR $\uparrow$        & 7.2    & 6.1    & 7.3    & 6.1    \\
    & Avg. Conf. (\%) $\uparrow$ & 88.0   & 94.4   & 85.7   & 94.2   \\
    \hline
    
    \multirow{3}{*}{Inc-v4}
    & SSIM $\uparrow$        & 0.0066 & 0.0049 & 0.0071 & 0.0046 \\
    & PSNR $\uparrow$        & 7.2    & 6.0    & 7.2    & 6.0    \\
    & Avg. Conf. (\%) $\uparrow$ & 93.4   & 95.8   & 88.9   & 86.5   \\
    \hline
    
    \multirow{3}{*}{Res-152}
    & SSIM $\uparrow$        & 0.0066 & 0.0046 & 0.0072 & 0.0048 \\
    & PSNR $\uparrow$        & 7.2    & 6.0    & 7.2    & 6.1    \\
    & Avg. Conf. (\%) $\uparrow$ & 94.3   & 93.3   & 88.7   & 89.3   \\
    \hline
    
    \multirow{3}{*}{IncRes-v2}
    & SSIM $\uparrow$        & 0.0064 & 0.0044 & 0.0068 & 0.0043 \\
    & PSNR $\uparrow$        & 7.2    & 6.0    & 7.1    & 6.0    \\
    & Avg. Conf. (\%) $\uparrow$ & 87.6   & 89.2   & 66.9   & 87.8   \\
    \hline
    \end{tabular}

    \begin{tabular}{l}
    \toprule[0pt]
    We present the similarity evaluation metrics between adversarial examples and original examples, which are generated for\\ Inc-v3, Inc-v4, IncRes-v2, and Res-152 using negI-FGSM, negI-FGM, negMI-FGM, and negMI-FGSM in a white-box setting.
    \end{tabular}

\end{table*}

Table \ref {table:2} shows the performance analysis of four different models under four adversarial attack methods, and all adversarial examples are generated in the white-box setting. To accurately measure image similarity, we use three key metrics: SSIM, PSNR, and average confidence rate. Higher SSIM and PSNR values indicate lower distortion and higher similarity between images. 

The experimental results show that the performance of the four adversarial attack methods in SSIM and PSNR indicators is relatively low, but the average confidence rate is high. Referring to the data in Table \ref{table:1}, we find that the accuracy of the model is also maintained at a high level in the white-box setting, indicating that although the adversarial examples are significantly different from the original image at the image level, their semantic information is still retained by the model. The model is still able to correctly identify the categories that these examples correspond to due to its high confidence and accuracy.

\begin{figure*}[!t]
\centering
\includegraphics[width=5.4in]{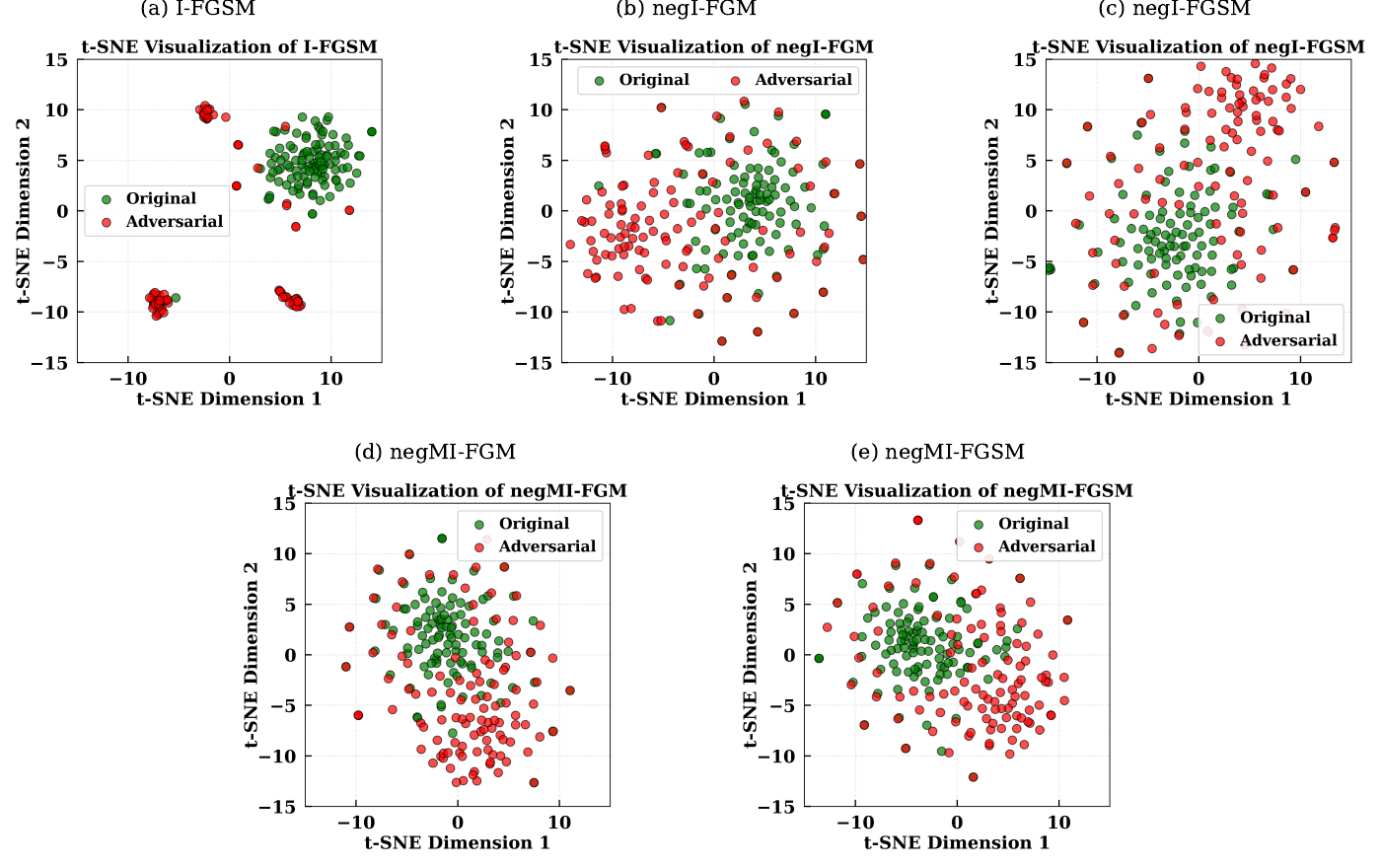}
\caption{Comparison of t-SNE distribution of high-level feature space under different adversarial attack methods}
\label{he}
\end{figure*}
Fig.\ref{he} shows the distribution characteristics of adversarial examples generated by the I-FGSM method and by the four methods of negI-FGM, negI-FGSM, negMI-FGM, and negMI-FGSM in the high-level feature space of the neural network, visualized with the t-SNE dimensionality-reduction technique. These adversarial examples are crafted for the Inc-v3 model in a white-box setting. The green scatter plot represents the feature distribution of the original examples, whereas the red scatter plot corresponds to that of the adversarial examples. The degree of overlap between the two in the feature space directly reflects how well adversarial attacks preserve the high-level semantic features of the original examples.

Specifically, the proposed methods accumulate historical gradient-direction information through a momentum mechanism, thereby preserving the core structure of the high-level semantic features more effectively. The experimental results reveal that the overlap achieved by the adversarial examples produced by the four proposed methods is significantly higher than that achieved by the examples generated by the I-FGSM method. This finding indicates that, although the adversarial examples created by our proposed attack methods may appear visually meaningless, they manage to protect the high-level semantic features of the original examples from being destroyed and still achieve correct classification.

\section{Conclusion}
In this paper, we reveal the inherent properties of neural networks. Specifically, the results show that the distribution of adversarial examples is extremely wide, extending not only to the neighbourhood of the data points but also to regions far from them. Therefore, the decision boundary should be appropriately contracted to exclude these outliers. In the future, we will explore the application of this feature in specific scenarios and also focus on developing more effective methods for generating the new type of adversarial examples.

\section*{Acknowledgments}
This work was supported by the Zhenjiang Jinshan Talent Program.

\bibliography{ref.bib}

\vfill

\end{document}